\documentclass[sigconf]{acmart}
\usepackage[mathscr]{eucal}
\usepackage{amsmath,amsthm,bm}
\usepackage{enumitem}
\usepackage{multirow}
\usepackage{booktabs}
\usepackage{scalerel}
\usepackage[multiple]{footmisc}
\usepackage{pgfplots}
\usepackage{pifont}
\usepackage{algorithm} 
\usepackage{algpseudocode}
\usepackage{tabularx}
\usepackage{soul}
\usepackage{makecell}
\usepackage{footnote}
\usepackage{balance}
\usepackage{graphicx}
\usepackage{nicefrac}
\usepackage{subcaption}
\usepackage{textcomp}
\usepackage{array}
\pgfplotsset{compat=1.18}

\makeatletter
\def\BibTeX{{\rm B\kern-.05em{\sc i\kern-.025em b}\kern-.08em
    T\kern-.1667em\lower.7ex\hbox{E}\kern-.125emX}}

\newcommand{\multiline}[1]{%
  \begin{tabularx}{\dimexpr\linewidth-\ALG@thistlm}[t]{@{}X@{}}
    #1
  \end{tabularx}
}

\def\addlegendimage{\csname pgfplots@addlegendimage\endcsname}
\usetikzlibrary{patterns}
\usetikzlibrary{pgfplots.groupplots}
\usetikzlibrary{
  pgfplots.colorbrewer,
}
\pgfplotsset{
  cycle list/.define={my marks}{
    every mark/.append style={solid,fill=\pgfkeysvalueof{/pgfplots/mark list fill}},mark=*\\
    every mark/.append style={solid,fill=\pgfkeysvalueof{/pgfplots/mark list fill}},mark=square*\\
    every mark/.append style={solid,fill=\pgfkeysvalueof{/pgfplots/mark list fill}},mark=triangle*\\
    every mark/.append style={solid,fill=\pgfkeysvalueof{/pgfplots/mark list fill}},mark=diamond*\\
  },
}

\DeclareMathSymbol{\mh}{\mathord}{operators}{`\-}
\definecolor{rank1}{HTML}{FFD966} 
\definecolor{rank2}{HTML}{C2F0C2} 
\newcommand{\hly}[1]{{\sethlcolor{rank1}\hl{#1}}} 
\newcommand{\hlb}[1]{{\sethlcolor{rank2}\hl{#1}}}   
\AtBeginDocument{%
  \providecommand\BibTeX{{%
    Bib\TeX}}}

\setcopyright{acmlicensed}
\copyrightyear{2026}
\acmYear{2026}
\setcopyright{cc}
\setcctype{by}
\acmConference[CIKM '26]{Proceedings of the 35th ACM International Conference on Information and Knowledge Management}{November 07--11, 2026}{Rome, Italy}
\acmBooktitle{Proceedings of the 35th ACM International Conference on Information and Knowledge Management (CIKM '26), November 07--11, 2026, Rome, Italy}
\acmDOI{10.1145/3799682.3839917}
\acmISBN{979-8-4007-2539-5/2026/11}

\newcommand{\ie}{{\it i.e.}}
\newcommand{\eg}{{\it e.g.}}
\newcommand{\cote}{{\textsc {\textsc CoTeach}}}

\begin{document}

\title{Who Should Teach? Confidence-Aware Dual-Teacher Learning for Few-Shot Node Classification on Text-Attributed Graphs} 

\author{Hojin Kim}
\affiliation{%
  \institution{Chungbuk National University}
  \city{Cheongju}
  \country{South Korea}
}
\email{khojin.01@cbnu.ac.kr}

\author{Sujin Yoon}
\affiliation{%
  \institution{Chungbuk National University}
  \city{Cheongju}
  \country{South Korea}
}
\email{sujin.yoon@cbnu.ac.kr}

\author{Sungsu Lim}
\affiliation{%
  \institution{Chungnam National University}
  \city{Daejeon}
  \country{South Korea}
}
\email{sungsu@cnu.ac.kr}

\author{Dongwon Lee}
\affiliation{%
 \institution{The Pennsylvania State University}
 \city{University Park, PA}
  \country{USA}
}
\email{dongwon@psu.edu}

\author{David Yoon Suk Kang}
\authornote{Corresponding author.}
\affiliation{%
  \institution{Chungbuk National University}
  \city{Cheongju}
  \country{South Korea}
}
\email{dyskang@cbnu.ac.kr}

\renewcommand{\shortauthors}{Hojin Kim, Sujin Yoon, Sungsu Lim, Dongwon Lee, \& David Yoon Suk Kang}

\begin{abstract}
Text-Attributed Graphs (TAGs) integrate graph structures and node-associated textual attributes, and recent studies have increasingly leveraged Large Language Models (LLMs) to improve TAG learning in few-shot settings. 
However, existing approaches typically utilize LLM-derived information uniformly across all nodes, despite substantial variations in its reliability, while also incurring considerable monetary costs. 
We argue that the most appropriate source of supervision may differ across nodes, as Graph Neural Networks (GNNs) and LLMs exhibit complementary strengths in exploiting structural and semantic information, respectively. 
To this end, we propose {\cote}, a \underline{\textbf{Co}}nfidence-aware dual-\underline{\textbf{teach}}er learning framework that dynamically selects the more reliable teacher for each node. Experimental results demonstrate that {\cote} consistently improves few-shot node classification performance while reducing unnecessary LLM utilization and associated monetary costs.

\end{abstract}




\maketitle

\section{Introduction}~\label{s1}

\vspace{-3mm}
\noindent \textbf{Background.} \textit{Text-Attributed Graphs} (TAGs) are graph-structured data that jointly encode relational information among nodes through graph topology and semantic information through node-associated textual attributes~\cite{jin23:acl, yan23:neurips, zha24:neurips}. 
Owing to their ability to capture both structural and semantic characteristics, TAGs have been widely employed in numerous real-world applications, including social networks, e-commerce platforms, and online communities~\cite{jin24:tkde}.
In these settings, effectively fusing graph structural information with node textual semantics is crucial for improving downstream graph learning tasks, such as node classification, link prediction, and recommendation~\cite{jin24:tkde}.

Existing TAG learning methods typically encode node texts using \textit{shallow textual representations} (\eg, TF-IDF~\cite{spa72:jd} and Word2Vec~\cite{mik13:arxiv}) or \textit{pretrained language models} (PLMs) (\eg, BERT~\cite{dev19:acl}) and integrate them with graph neural networks (GNNs)~\cite{kip17:iclr, vel17:arxiv, ham17:neurips} for graph learning~\cite{jin23:acl, yan23:neurips}.
However, these methods largely treat textual semantics as \textit{static} representations and cannot fully exploit the rich linguistic knowledge encoded in language models~\cite{he24:iclr, ji24:arxiv}.

\begin{figure*}[t]
\centering
\includegraphics[width=0.85\textwidth]{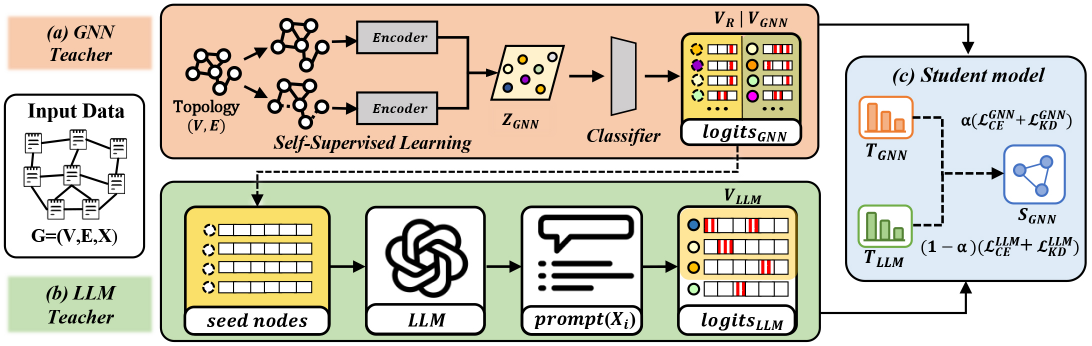}
\vspace{-4mm}
\caption{Overview of TAG learning in {\cote}.} \label{fig_1}
\vspace{-5mm}
\end{figure*}

To address this limitation, recent studies have begun incorporating Large Language Models (LLMs) directly into TAG learning frameworks~\cite{ren24:kdd, jin24:tkde}. 
Representative paradigms include \textit{LLM-as-Annotator}~\cite{che24:iclr}, \textit{LLM-as-Enhancer}~\cite{he24:iclr}, and \textit{LLM-as-Generator}~\cite{yu25:aaai}, which leverage LLMs for annotation, representation enhancement, and data generation, respectively.
These approaches have demonstrated promising improvements, particularly in label-scarce and few-shot settings~\cite{che24:iclr, he24:iclr, yu25:aaai}. 

\noindent \textbf{Motivation.} Existing LLM-based TAG learning methods leverage LLM-derived supervision to enhance graph representation learning~\cite{che24:iclr, he24:iclr, yu25:aaai}. 
However, they generally do not explicitly account for variations in the reliability of LLM supervision across nodes.
In practice, the quality of LLM-generated supervision can vary substantially across nodes. 
While LLM outputs may provide highly informative guidance for certain nodes, they can introduce noisy or incorrect supervision for others.
For example, generated pseudo-labels may contain errors for specific nodes, whereas augmented semantic information or explanations may not be equally beneficial across the graph~\cite{he24:iclr, che24:iclr}.

However, LLMs are not the only source of supervision available in TAGs.
Graph topology itself provides valuable structural signals, and GNN-based approaches have demonstrated that neighboring nodes can effectively guide representation learning through message passing~\cite{kip17:iclr, vel17:arxiv, ham17:neurips}.
Consequently, different nodes may benefit from different sources of supervision.
While some nodes may be better guided by semantic knowledge from LLMs, others may benefit more from structural signals encoded in graph topology.
This raises a fundamental question:
\textit{Which source of supervision should guide each node?}
In other words, \textit{Who should teach each node?}

\vspace{1mm}
\noindent \textbf{Proposed Framework.} To answer this question, we propose {\cote}, a confidence-aware dual-teacher learning framework for few-shot TAG learning. 
{\cote} incorporates two complementary teachers: a \textit{GNN Teacher} that captures structural knowledge from graph topology and an \textit{LLM Teacher} that exploits semantic knowledge from node texts. 
For each node, {\cote} first evaluates the confidence of GNN predictions. Nodes with high-confidence predictions are directly supervised by the GNN teacher, whereas uncertain nodes are further examined by the LLM teacher. 
Only reliable LLM predictions, as determined by confidence scoring, are incorporated as supervision.
Through this confidence-aware teacher selection, {\cote} provides a simple yet effective mechanism for adaptive supervision in few-shot TAG learning while reducing unnecessary monetary costs associated with commercial LLM inference.

\vspace{1mm}
\noindent \textbf{Contributions.}
The contributions of this paper are as follows:
\begin{itemize}[leftmargin=*]
    \item \textbf{New Perspective.} We introduce an adaptive supervision perspective that accounts for node-level variations in the reliability of LLM-derived supervision.
    \item \textbf{Novel Framework.} We propose {\cote}, a confidence-aware dual-teacher learning framework that adaptively selects the most reliable teacher for each node.
    \item \textbf{Extensive Evaluation.} We conduct comprehensive evaluations on four real-world TAGs, validating the effectiveness and cost efficiency of {\cote}.
\end{itemize}

\section{{\cote}: Proposed Framework}~\label{s4}

\vspace{-3mm}
\noindent \textbf{Overview.} In this section, we present {\cote}, a confidence-aware dual-teacher learning framework for few-shot text-attributed graph learning.
As illustrated in Figure~\ref{fig_1}, {\cote} consists of three main components: \textbf{(S1)} GNN teacher, \textbf{(S2)} LLM teacher, and \textbf{(S3)} Student model.
Given a TAG, the GNN teacher first generates pseudo-labels and confidence scores based on structural information (Figure~\ref{fig_1}-(a)). 
Nodes with high-confidence predictions are directly supervised by the GNN teacher, whereas nodes with low-confidence predictions are delegated to the LLM teacher for semantic reasoning. 
The LLM teacher then produces complementary pseudo-labels for these uncertain nodes using textual information (Figure~\ref{fig_1}-(b)). 
Finally, the student model learns from confidence-aware pseudo-labels provided by both teachers (Figure~\ref{fig_1}-(c)).


\vspace{1mm}
\noindent \textbf{GNN Teacher.} 
The \textit{GNN teacher} is designed to generate structure-aware supervision by exploiting relational information embedded in the graph topology. 
Specifically, we employ Deep Graph Infomax (DGI)~\cite{ve18:iclr} to learn node representations in a self-supervised manner and train a two-layer MLP classifier~\cite{ga98:ae} on top of the learned embeddings.
The GNN Teacher then produces a pseudo-label and a confidence score for each node, where the pseudo-label corresponds to the most probable class and the confidence score is defined as its associated prediction probability.
Using a confidence threshold $\gamma_g$, each node $v_i$ in $\mathcal{V}$ are partitioned into two groups:
\vspace{-1mm}
\begin{equation}
\mathcal{V}_{GNN}
=
\{v_i \in \mathcal{V}\mid c_i^g \ge \gamma_g\},
\qquad
\mathcal{V}_R
=
\mathcal{V}\setminus\mathcal{V}_{GNN}.
\end{equation}
Here, $\mathcal{V}_{GNN}$ denotes the set of nodes for which the GNN teacher makes highly confident predictions, while $\mathcal{V}_R$ denotes the set of nodes with relatively low confidence. 
We interpret nodes in $\mathcal{V}_R$ as cases where structural information alone is insufficient to generate reliable predictions.
Consequently, the GNN teacher provides reliable pseudo-labels for structurally informative nodes while identifying nodes for which structural information alone is insufficient for confident prediction. 
These nodes are subsequently passed to the LLM teacher, which provides complementary supervision based on semantic information.


\vspace{1mm}
\noindent \textbf{LLM Teacher.} 
The \textit{LLM teacher} is designed to generate semantic-aware supervision by leveraging the textual semantics of nodes. It focuses on nodes in $\mathcal{V}_R$, where structural information alone is insufficient for reliable prediction.
Since querying LLMs for all nodes in $\mathcal{V}_R$ incurs substantial monetary costs, we first select a subset of representative nodes using a seed node selection strategy.
Specifically, following~\cite{zha25:sigir}, we cluster node representations and select nodes closest to the cluster centroids as seed nodes. 
This strategy reduces the number of LLM queries while selecting semantically representative nodes from diverse regions of the feature space, thereby encouraging balanced semantic supervision across different classes.

For each selected node, the LLM teacher receives a prompt consisting of the node text and candidate label descriptions, and predicts a probability distribution over the candidate labels.\footnote{An example prompt is available at: https://buly.kr/GP4yN34}
The pseudo-label is assigned as the label with the highest predicted probability, while the corresponding probability is used as the confidence score.
Only predictions whose confidence scores exceed a predefined threshold $\gamma_l$ are retained and assigned to the node set $\mathcal{V}_{LLM}$, while the remaining predictions are discarded.
The resulting node set $\mathcal{V}_{LLM}$ is then used, together with $\mathcal{V}_{GNN}$, to supervise the student model.

\vspace{1mm}
\noindent \textbf{Student Model.}
The \textit{Student model} is a component designed to learn node representations under the supervision of both the GNN teacher and the LLM teacher.
Specifically, nodes in $\mathcal{V}_{GNN}$ are supervised using the pseudo-labels and class probability distributions produced by the GNN teacher, whereas nodes in $\mathcal{V}_{LLM}$ are supervised using those generated by the LLM teacher. 
In this way, each node selectively receives supervision from the teacher that provides the most reliable prediction.
The student produces node-level class logits as follows:
\vspace{-2mm}
\begin{equation}
\mathbf{z}_i^s
=
g_{\phi_s}
\left(
f_{\theta_s}(\mathbf{X},\mathbf{A})
\right),
\end{equation}
where $f_{\theta_s}$ and $g_{\phi_s}$ denote the student GNN encoder and classification head, respectively.
The student is trained using both hard pseudo-label supervision and soft distribution distillation. 
To learn the pseudo-labels generated by each teacher, we employ the following cross-entropy loss:

\begin{equation}
\mathcal{L}_{CE}^{T}
=
\sum_{v_i\in\mathcal{V}_{T}}
\mathrm{CE}
\left(
\mathbf{z}_i^s,
\hat{y}_i^{T}
\right),
\end{equation}
where $T\in\{GNN,LLM\}$, $\mathcal{V}_{T}$ denotes the node set supervised by teacher $T$, and $\hat{y}_i^{T}$ denotes the corresponding pseudo-label.
To further transfer the uncertainty and inter-class relationships captured by each teacher, we perform knowledge distillation by matching the class probability distributions:
\begin{equation}
\mathcal{L}_{KD}^{T}
=
\sum_{v_i\in\mathcal{V}_{T}}
\mathrm{KL}
\left(
\mathbf{p}_i^{T}
\Vert
\mathbf{p}_i^{s}
\right),
\end{equation}
where $\mathbf{p}_i^{T}$ and $\mathbf{p}_i^{s}$ denote the class probability distributions of the teacher and the student, respectively.
The final training objective is defined as
\begin{equation}
\mathcal{L}
=
\alpha
\left(
\mathcal{L}_{CE}^{GNN}
+
\mathcal{L}_{KD}^{GNN}
\right)
+
(1-\alpha)
\left(
\mathcal{L}_{CE}^{LLM}
+
\mathcal{L}_{KD}^{LLM}
\right),
\end{equation}
where $\alpha \in [0,1]$ controls the relative contributions of the GNN teacher and LLM teacher.
Consequently, the student effectively combines structure-aware supervision from the GNN teacher and semantic supervision from the LLM teacher, leading to improved few-shot node classification performance.

\section{Evaluation}~\label{s5}
We evaluate {\cote} to answer the following questions:

\begin{itemize}[leftmargin=*]
    \item \textbf{EQ1.} How does each teacher contribute to the effectiveness of {\cote}?
    \item \textbf{EQ2.}: Does {\cote} outperform existing state-of-the-art TAG learning methods?
    \item \textbf{EQ3.} How cost-efficient is {\cote} compared with existing LLM-based methods?
    \item \textbf{EQ4.} How sensitive is {\cote} to key hyperparameters?
\end{itemize}

\subsection{Experimental Settings}
\noindent \textbf{Datasets and TAG Learning Methods.}
Following~\cite{yu25:aaai, li24:bd, zha25:sigir}, we evaluate \textsf{\cote} on 4 real-world TAGs: Cora, Citeseer, Pubmed, and WikiCS.
Here, we summarize the basic statistics of these datasets in Table~\ref{tab_3}, where |$V$|, |$E$|, and |$C$| denote the number of nodes, edges, and classes, respectively.
We compare {\cote} with the following 10 competing methods: (1) 3 backbone GNN methods (\ie, GCN~\cite{kip17:iclr}, GAT~\cite{vel17:arxiv}, and GraphSAGE~\cite{ham17:neurips}); (2) 3 PLM methods (\ie, BERT~\cite{dev19:acl}, BART~\cite{lew20:acl}, and SBERT~\cite{rei19:emnlp}); (3) 1 GNN+PLM method (\ie, GLEM~\cite{zha23:iclr}); and (4) 3 GNN+LLM methods (\ie, LLMGNN~\cite{che24:iclr}, TAPE~\cite{he24:iclr}, and LLM4NG~\cite{yu25:aaai}). 
All hyperparameters are configured using the best settings identified through an extensive grid search within the ranges recommended in the corresponding papers.

\vspace{1mm}
\noindent \textbf{Implementation.}
In {\cote}, GPT-3.5-turbo is used as the LLM Teacher, while GCN, GAT, and GraphSAGE serve as the student model backbones. 
To facilitate reproducibility, we publicly release the source code, hyperparameter settings, and implementation details at: https://buly.kr/GP4yN34.

\vspace{1mm}
\noindent \textbf{Evaluation Protocol.} 
Following~\cite{li24:bd}, we split each dataset into training, validation, and test sets with a ratio of 60\%, 20\%, and 20\%, respectively. 
For $k$-shot learning, we randomly sample $k \in \{3,5,7,10\}$ labeled nodes per class from the training set. 
Performance is evaluated by test accuracy and reported as the mean and standard deviation over five runs.
\begin{table}[t] 
\centering 
\footnotesize 
\caption{Statistics of real-world TAGs used in our experiments} \label{tab_3} 
\vspace{-4mm} 
\def\arraystretch{1.05} 
\setlength{\tabcolsep}{4.5pt} 
\begin{tabular}{c|ccccc} 
\toprule  
& 
|$V$| & |$E$| & |$C$| & Domain & Text Attribute
 \\ \midrule 
\midrule 
\textbf{Cora} & 2,078 & 5,278 & 7  &Citation& Title \& abstract of a paper\\ 
\textbf{Citeseer} & 3,186 & 4,225 & 6 &Citation& Title \& abstract of a paper \\ 
\textbf{Pubmed} & 19,717 & 44,324 & 3 &Citation& Title \& abstract of a paper \\ 
\textbf{WikiCS} & 11,701 & 215,603 & 10 & Web & Content of a article \\ 
\bottomrule 
\end{tabular} 
\vspace{-5mm} 
\end{table}

\subsection{Experimental Results}


\begin{table*}[t]
\centering
\footnotesize
\caption{Few-shot node classification accuracy of {\cote} and 10 competing methods on four real-world datasets.}
\label{tab_1}
\vspace{-4mm}
\def\arraystretch{1.05}
\setlength{\tabcolsep}{1.5pt}
\resizebox{\textwidth}{!}{
\begin{tabular}{ll|ccc|ccc|c|ccc|ccc|c}
\toprule

& &
\multicolumn{3}{c|}{\textbf{GNN}}
&
\multicolumn{3}{c|}{\textbf{PLM}}
&
\multicolumn{1}{c|}{\textbf{GNN+PLM}}
&
\multicolumn{6}{c|}{\textbf{GNN+LLM}}
& \multirow{2}{*}{\makecell{\textbf{Gain}\\(\%)}}
\\

\cmidrule(lr){3-5}
\cmidrule(lr){6-8}
\cmidrule(lr){9-9}
\cmidrule(lr){10-15}
 & \textbf{k}
& \textbf{GCN}
& \textbf{GAT}
& \textbf{GraphSAGE}
& \textbf{BERT}
& \textbf{BART}
& \textbf{SBERT}
& \textbf{GLEM}
& \textbf{TAPE}
& \textbf{LLMGNN}
& \textbf{LLM4NG}
& \textsc{\textbf{CoTeach}}$_{\textbf{G}}$
& \textsc{\textbf{CoTeach}}$_{\textbf{A}}$
& \textsc{\textbf{CoTeach}}$_{\textbf{S}}$
& 
\\
\midrule
\midrule
\multirow{4}{*}{\rotatebox[origin=c]{90}{\textbf{Cora}}}

& \textbf{3}
& 65.5$\pm$2.7 {\scriptsize(7)}
& 64.3$\pm$4.7 {\scriptsize(8)}
& 59.0$\pm$4.3 {\scriptsize(9)}
& 27.4$\pm$3.5 {\scriptsize(13)}
& 28.8$\pm$4.8 {\scriptsize(12)}
& 50.6$\pm$3.6 {\scriptsize(11)}
& 52.8$\pm$6.8 {\scriptsize(10)}
& 65.6$\pm$4.1 {\scriptsize(6)}
& 68.6$\pm$2.5 {\scriptsize(5)}
& \hlb{\mbox{\underline{72.2$\pm$3.8 {\scriptsize(4)}}}}
& 76.5$\pm$2.4 {\scriptsize(2)}
& \hly{\textbf{77.5$\pm$2.9 {\scriptsize(1)}}}
& 75.7$\pm$2.5 {\scriptsize(3)}
& \textcolor{blue}{+7.3}
\\

& \textbf{5}
& 73.0$\pm$3.7 {\scriptsize(6)}
& 71.8$\pm$4.6 {\scriptsize(7)}
& 68.5$\pm$1.3 {\scriptsize(9)}
& 30.1$\pm$4.6 {\scriptsize(13)}
& 31.6$\pm$6.4 {\scriptsize(12)}
& 56.6$\pm$3.1 {\scriptsize(11)}
& 59.1$\pm$5.5 {\scriptsize(10)}
& 73.6$\pm$3.5 {\scriptsize(5)}
& 70.4$\pm$2.1 {\scriptsize(8)}
& \hlb{\mbox{\underline{76.8$\pm$2.6 {\scriptsize(4)}}}}
& \hly{\textbf{81.0$\pm$1.1 {\scriptsize(1)}}}
& 81.0$\pm$1.4 {\scriptsize(2)}
& 80.2$\pm$0.8 {\scriptsize(3)}
& \textcolor{blue}{+5.5}
\\

& \textbf{7}
& 74.5$\pm$4.7 {\scriptsize(6)}
& 74.8$\pm$2.8 {\scriptsize(5)}
& 72.6$\pm$1.9 {\scriptsize(8)}
& 32.2$\pm$5.3 {\scriptsize(13)}
& 33.2$\pm$5.5 {\scriptsize(12)}
& 58.8$\pm$3.6 {\scriptsize(11)}
& 66.7$\pm$3.1 {\scriptsize(10)}
& 74.0$\pm$2.7 {\scriptsize(7)}
& 71.8$\pm$2.3 {\scriptsize(9)}
& \hlb{\mbox{\underline{78.6$\pm$1.1 {\scriptsize(4)}}}}
& 81.5$\pm$1.0 {\scriptsize(2)}
& \hly{\textbf{82.0$\pm$0.5 {\scriptsize(1)}}}
& 81.3$\pm$1.5 {\scriptsize(3)}
& \textcolor{blue}{+4.3}
\\

& \textbf{10}
& 76.7$\pm$2.9 {\scriptsize(5)}
& 75.1$\pm$3.8 {\scriptsize(7)}
& 74.6$\pm$1.4 {\scriptsize(8)}
& 34.9$\pm$3.8 {\scriptsize(13)}
& 35.1$\pm$5.5 {\scriptsize(12)}
& 60.9$\pm$2.9 {\scriptsize(11)}
& 71.0$\pm$4.1 {\scriptsize(10)}
& 76.2$\pm$2.6 {\scriptsize(6)}
& 72.9$\pm$2.1 {\scriptsize(9)}
& \hlb{\mbox{\underline{80.7$\pm$1.5 {\scriptsize(4)}}}}
& 82.1$\pm$1.4 {\scriptsize(3)}
& \hly{\textbf{82.8$\pm$2.3 {\scriptsize(1)}}}
& 82.4$\pm$1.5 {\scriptsize(2)}
& \textcolor{blue}{+2.6}
\\
\midrule
\multirow{4}{*}{\rotatebox[origin=c]{90}{\textbf{Citeseer}}}

& \textbf{3}
& 46.9$\pm$6.9 {\scriptsize(10)}
& 49.2$\pm$3.5 {\scriptsize(8)}
& 43.0$\pm$3.6 {\scriptsize(11)}
& 31.8$\pm$3.6 {\scriptsize(13)}
& 33.8$\pm$4.2 {\scriptsize(12)}
& 55.2$\pm$4.0 {\scriptsize(7)}
& 48.8$\pm$6.9 {\scriptsize(9)}
& 62.3$\pm$3.4 {\scriptsize(5)}
& 61.3$\pm$3.2 {\scriptsize(6)}
& \hlb{\mbox{\underline{67.1$\pm$1.6 {\scriptsize(3)}}}}
& 65.1$\pm$1.9 {\scriptsize(4)}
& 67.6$\pm$1.2 {\scriptsize(2)}
& \hly{\textbf{68.1$\pm$1.8 {\scriptsize(1)}}}
& \textcolor{blue}{+1.5}
\\

& \textbf{5}
& 51.2$\pm$4.0 {\scriptsize(10)}
& 55.3$\pm$4.7 {\scriptsize(8)}
& 47.2$\pm$7.5 {\scriptsize(11)}
& 33.3$\pm$3.7 {\scriptsize(13)}
& 36.3$\pm$5.5 {\scriptsize(12)}
& 58.4$\pm$3.3 {\scriptsize(7)}
& 54.6$\pm$3.6 {\scriptsize(9)}
& 63.9$\pm$5.1 {\scriptsize(5)}
& 62.4$\pm$1.9 {\scriptsize(6)}
& \hlb{\mbox{\underline{67.2$\pm$1.8 {\scriptsize(3)}}}}
& 66.8$\pm$1.8 {\scriptsize(4)}
& \hly{\textbf{69.1$\pm$2.1 {\scriptsize(1)}}}
& 68.5$\pm$3.0 {\scriptsize(2)}
& \textcolor{blue}{+2.8}
\\

& \textbf{7}
& 57.3$\pm$4.2 {\scriptsize(10)}
& 59.6$\pm$3.1 {\scriptsize(8)}
& 53.5$\pm$5.4 {\scriptsize(11)}
& 33.6$\pm$1.1 {\scriptsize(13)}
& 38.8$\pm$4.0 {\scriptsize(12)}
& 58.1$\pm$2.2 {\scriptsize(9)}
& 60.7$\pm$6.5 {\scriptsize(7)}
& 65.8$\pm$4.1 {\scriptsize(5)}
& 63.0$\pm$1.9 {\scriptsize(6)}
& \hlb{\mbox{\underline{67.9$\pm$1.3 {\scriptsize(3)}}}}
& 67.2$\pm$3.3 {\scriptsize(4)}
& 68.8$\pm$2.9 {\scriptsize(2)}
& \hly{\textbf{69.3$\pm$1.3 {\scriptsize(1)}}}
& \textcolor{blue}{+2.1}
\\

& \textbf{10}
& 58.8$\pm$3.3 {\scriptsize(10)}
& 60.0$\pm$2.4 {\scriptsize(8)}
& 56.0$\pm$5.2 {\scriptsize(11)}
& 36.0$\pm$1.3 {\scriptsize(13)}
& 42.9$\pm$3.4 {\scriptsize(12)}
& 59.8$\pm$3.4 {\scriptsize(9)}
& 64.5$\pm$2.4 {\scriptsize(6)}
& 66.3$\pm$3.6 {\scriptsize(5)}
& 63.3$\pm$2.1 {\scriptsize(7)}
& \hlb{\mbox{\underline{68.3$\pm$1.1 {\scriptsize(3)}}}}
& 67.9$\pm$2.7 {\scriptsize(4)}
& \hly{\textbf{70.3$\pm$1.4 {\scriptsize(1)}}}
& 69.8$\pm$2.8 {\scriptsize(2)}
& \textcolor{blue}{+2.9}
\\
\midrule
\multirow{4}{*}{\rotatebox[origin=c]{90}{\textbf{Pubmed}}}

& \textbf{3}
& 63.6$\pm$8.8 {\scriptsize(7)}
& 62.6$\pm$8.5 {\scriptsize(8)}
& 58.9$\pm$6.9 {\scriptsize(9)}
& 36.8$\pm$3.3 {\scriptsize(13)}
& 55.8$\pm$7.2 {\scriptsize(11)}
& 44.9$\pm$7.7 {\scriptsize(12)}
& 56.4$\pm$5.1 {\scriptsize(10)}
& 72.2$\pm$9.8 {\scriptsize(6)}
& \hlb{\mbox{\underline{79.7$\pm$1.3 {\scriptsize(4)}}}}
& 74.9$\pm$3.7 {\scriptsize(5)}
& 80.6$\pm$1.3 {\scriptsize(3)}
& 81.5$\pm$2.7 {\scriptsize(2)}
& \hly{\textbf{81.9$\pm$2.5 {\scriptsize(1)}}}
& \textcolor{blue}{+2.8}
\\

& \textbf{5}
& 69.8$\pm$5.7 {\scriptsize(7)}
& 69.5$\pm$6.7 {\scriptsize(8)}
& 66.4$\pm$7.0 {\scriptsize(9)}
& 38.3$\pm$2.4 {\scriptsize(13)}
& 57.5$\pm$9.0 {\scriptsize(11)}
& 50.9$\pm$5.0 {\scriptsize(12)}
& 59.0$\pm$3.6 {\scriptsize(10)}
& 79.0$\pm$4.5 {\scriptsize(5)}
& \hlb{\mbox{\underline{81.6$\pm$1.2 {\scriptsize(3)}}}}
& 77.4$\pm$3.1 {\scriptsize(6)}
& 81.2$\pm$3.2 {\scriptsize(4)}
& 81.9$\pm$1.7 {\scriptsize(2)}
& \hly{\textbf{82.2$\pm$1.7 {\scriptsize(1)}}}
& \textcolor{blue}{+0.7}
\\

& \textbf{7}
& 72.7$\pm$5.8 {\scriptsize(7)}
& 71.9$\pm$6.2 {\scriptsize(8)}
& 71.3$\pm$4.6 {\scriptsize(9)}
& 39.4$\pm$5.5 {\scriptsize(13)}
& 59.6$\pm$6.4 {\scriptsize(11)}
& 51.3$\pm$6.2 {\scriptsize(12)}
& 63.0$\pm$3.1 {\scriptsize(10)}
& 79.7$\pm$3.3 {\scriptsize(5)}
& \hlb{\mbox{\underline{82.4$\pm$1.2 {\scriptsize(4)}}}}
& 78.6$\pm$2.7 {\scriptsize(6)}
& 82.8$\pm$0.9 {\scriptsize(3)}
& 83.0$\pm$0.7 {\scriptsize(2)}
& \hly{\textbf{83.2$\pm$1.4 {\scriptsize(1)}}}
& \textcolor{blue}{+1.0}
\\

& \textbf{10}
& 76.2$\pm$2.3 {\scriptsize(7)}
& 75.4$\pm$2.2 {\scriptsize(8)}
& 74.3$\pm$1.9 {\scriptsize(9)}
& 40.2$\pm$5.2 {\scriptsize(13)}
& 62.4$\pm$6.0 {\scriptsize(11)}
& 54.2$\pm$4.5 {\scriptsize(12)}
& 63.4$\pm$4.8 {\scriptsize(10)}
& 81.6$\pm$1.9 {\scriptsize(5)}
& \hlb{\mbox{\underline{83.0$\pm$1.0 {\scriptsize(3)}}}}
& 80.4$\pm$1.4 {\scriptsize(6)}
& 83.5$\pm$1.3 {\scriptsize(2)}
& 82.2$\pm$1.4 {\scriptsize(4)}
& \hly{\textbf{83.6$\pm$2.0 {\scriptsize(1)}}}
& \textcolor{blue}{+0.7}
\\
\midrule
\multirow{4}{*}{\rotatebox[origin=c]{90}{\textbf{WikiCS}}}

& \textbf{3}
& 57.3$\pm$6.5 {\scriptsize(7)}
& 58.7$\pm$4.6 {\scriptsize(6)}
& 55.6$\pm$3.8 {\scriptsize(9)}
& 30.5$\pm$4.5 {\scriptsize(13)}
& 36.4$\pm$1.3 {\scriptsize(12)}
& 43.4$\pm$6.5 {\scriptsize(11)}
& 51.9$\pm$9.7 {\scriptsize(10)}
& 56.1$\pm$4.1 {\scriptsize(8)}
& 71.7$\pm$2.4 {\scriptsize(5)}
& \hlb{\mbox{\underline{72.1$\pm$4.3 {\scriptsize(4)}}}}
& 72.2$\pm$4.1 {\scriptsize(3)}
& \hly{\textbf{74.2$\pm$3.6 {\scriptsize(1)}}}
& 74.0$\pm$2.3 {\scriptsize(2)}
& \textcolor{blue}{+2.9}
\\

& \textbf{5}
& 65.2$\pm$3.9 {\scriptsize(6)}
& 63.2$\pm$5.6 {\scriptsize(7)}
& 62.4$\pm$3.9 {\scriptsize(9)}
& 35.9$\pm$3.4 {\scriptsize(13)}
& 41.9$\pm$2.4 {\scriptsize(12)}
& 46.8$\pm$5.1 {\scriptsize(11)}
& 62.8$\pm$9.5 {\scriptsize(8)}
& 61.1$\pm$3.8 {\scriptsize(10)}
& 72.4$\pm$1.5 {\scriptsize(5)}
& \hlb{\mbox{\underline{72.5$\pm$3.9 {\scriptsize(4)}}}}
& 73.8$\pm$2.7 {\scriptsize(3)}
& \hly{\textbf{76.2$\pm$2.0 {\scriptsize(1)}}}
& 75.3$\pm$2.1 {\scriptsize(2)}
& \textcolor{blue}{+5.1}
\\

& \textbf{7}
& 67.4$\pm$2.8 {\scriptsize(7)}
& 65.2$\pm$3.9 {\scriptsize(8)}
& 64.8$\pm$3.6 {\scriptsize(9)}
& 40.4$\pm$1.8 {\scriptsize(13)}
& 44.1$\pm$1.9 {\scriptsize(12)}
& 50.4$\pm$3.6 {\scriptsize(11)}
& 68.0$\pm$7.5 {\scriptsize(6)}
& 64.0$\pm$6.1 {\scriptsize(10)}
& 71.9$\pm$1.2 {\scriptsize(5)}
& \hlb{\mbox{\underline{72.9$\pm$3.3 {\scriptsize(4)}}}}
& 74.6$\pm$2.8 {\scriptsize(3)}
& \hly{\textbf{76.0$\pm$1.6 {\scriptsize(1)}}}
& 75.6$\pm$1.8 {\scriptsize(2)}
& \textcolor{blue}{+4.3}
\\

& \textbf{10}
& 70.0$\pm$2.6 {\scriptsize(7)}
& 68.2$\pm$4.3 {\scriptsize(9)}
& 68.9$\pm$3.2 {\scriptsize(8)}
& 44.0$\pm$3.2 {\scriptsize(13)}
& 47.7$\pm$2.9 {\scriptsize(12)}
& 53.8$\pm$3.1 {\scriptsize(11)}
& 72.2$\pm$4.3 {\scriptsize(6)}
& 66.9$\pm$4.5 {\scriptsize(10)}
& 73.5$\pm$0.9 {\scriptsize(5)}
& \hlb{\mbox{\underline{75.2$\pm$1.7 {\scriptsize(4)}}}}
& 76.3$\pm$1.4 {\scriptsize(3)}
& \hly{\textbf{77.4$\pm$1.2 {\scriptsize(1)}}}
& 77.4$\pm$2.1 {\scriptsize(2)}
& \textcolor{blue}{+2.9}
\\
\midrule
\midrule
\multicolumn{2}{c|}{\textbf{AR $\downarrow$}}
& 7.4
& 7.6
& 9.3
& 13.0
& 11.8
& 10.5
& 8.8
& 6.4
& 5.4
& 4.2
& 3.0
& 1.6
& \textbf{1.8}
& --
\\
\bottomrule
\end{tabular}
}
\vspace{-5mm}
\end{table*}

\vspace{1mm}
\noindent \textbf{EQ1: Ablation Study.} We first individually assess the effectiveness of the supervision strategies proposed in {\cote}. 
Specifically, we compare four settings: no teacher, GNN teacher only, LLM teacher only, and both GNN and LLM teachers. 
For a fair comparison, all settings employ GraphSAGE as the student backbone model.
Note that the no teacher setting corresponds to the original GraphSAGE.
Unless otherwise specified, the results in this section are reported on WikiCS using GraphSAGE as the student backbone model due to space limitations. 
Similar trends were observed across the other datasets and backbone architectures.

We summarize the results in Figure~\ref{fig_2} as follows.
First, combining both GNN and LLM teachers consistently achieves the best performance across all few-shot settings. 
This result demonstrates that structural knowledge from the GNN teacher and semantic knowledge from the LLM teacher are complementary and can be effectively integrated to provide more reliable supervision.
Second, employing either the GNN teacher or the LLM teacher alone consistently outperforms the no teacher setting. 
This observation confirms the effectiveness of teacher-generated supervision for improving node classification performance under limited labeled data.
Third, neither the GNN teacher nor the LLM teacher consistently dominates the other across different few-shot settings. 
This suggests that the two teachers capture different yet complementary aspects of node information, motivating the need to jointly exploit both supervision sources.

\vspace{1mm}
\noindent \textbf{EQ2: Accuracy Comparison.} 
Next, we evaluate the performance of {\cote} against 10 competing methods. 
Table~\ref{tab_1} shows the results for k-shot node classification. 
{For each dataset, the best performance among \textit{{\cote} variants} and the best performance among \textit{existing competing methods} are highlighted in \hly{\textbf{bold}} and \hlb{\mbox{\underline{underline}}}, respectively.
\textit{AR} denotes the \textit{average rank} across all datasets. 
{\cote}$_{G}$, {\cote}$_{A}$, and {\cote}$_{S}$ denote the variants of {\cote} employing GCN, GAT, and GraphSAGE as the backbone architecture of the student model, respectively.

We summarize the results in Table~\ref{tab_1} as follows.
First, {\cote} achieves the best overall performance across most settings.
In particular, {\cote} attains the best overall average rank, with {\cote}$_S$ achieving an AR of 1.6, substantially outperforming existing baselines. 
These results demonstrate the effectiveness of the proposed confidence-aware dual-teacher framework for few-shot TAG learning.
Second, {\cote} consistently improves over strong baselines across most few-shot settings, with particularly notable gains on Cora and Pubmed. 
For instance, under the most challenging setting ($k=3$), {\cote} outperforms the strongest baseline by 6.9\% and 4.9\% on Cora and Pubmed, respectively. 
These results indicate that adaptive teacher selection can effectively improve pseudo-label quality, especially when reliable supervision is limited.
Third, {\cote} remains effective across different student backbone architectures. 
Although GCN, GAT, and GraphSAGE exhibit different learning characteristics, all {\cote} variants consistently rank among the top-performing methods. 
While certain variants occasionally underperform the strongest baseline, all variants remain competitive, and at least one {\cote} variant achieves the best performance in every dataset and few-shot setting. 
Combined with the findings of \textbf{EQ1}, these results further confirm the effectiveness of confidence-aware teacher selection.
\begin{figure}[t]
\footnotesize
\centering

\begin{tikzpicture}
\begin{axis}[
    ybar,
    width=1.05\linewidth,
    height=2.5cm,
    bar width=7pt,
    ylabel={Accuracy (\%)},
    symbolic x coords={3,5,7,10},
    xtick=data,
    ymin=40,
    ymax=80,
    enlarge x limits=0.15,
    grid=major,
    grid style={gray!20,dashed},
    legend style={
        at={(0.5,1.02)},
        anchor=south,
        legend columns=4,
        draw=none,
        font=\footnotesize,
        /tikz/every even column/.append style={column sep=0.2cm}
    },
    tick label style={font=\footnotesize},
    label style={font=\footnotesize},
]

\addplot[
    fill=gray!30,
    draw=gray!60,
    fill opacity=0.9
]
coordinates {
    (3,55.6)
    (5,62.4)
    (7,64.8)
    (10,68.9)
};
\addlegendentry{No Teacher}

\addplot[
    fill=purple!50,
    draw=purple!80,
    fill opacity=0.9
]
coordinates {
    (3,65.5)
    (5,67.5)
    (7,69.4)
    (10,72.9)
};
\addlegendentry{GNN Only}

\addplot[
    fill=teal!60,
    draw=teal!90,
    fill opacity=0.9
]
coordinates {
    (3,68.8)
    (5,68.7)
    (7,68.7)
    (10,68.7)
};
\addlegendentry{LLM Only}

\addplot[
    fill=orange!85,
    draw=orange!95!black,
    fill opacity=0.9,
    line width=0.8pt
]
coordinates {
    (3,73.8)
    (5,74.6)
    (7,75.3)
    (10,76.8)
};
\addlegendentry{GNN+LLM}

\end{axis}
\end{tikzpicture}
\vspace{-5mm}
\caption{
Comparison of different supervision strategies on WikiCS under varying few-shot settings.
}

\vspace{-6mm}
\label{fig_2}
\end{figure}
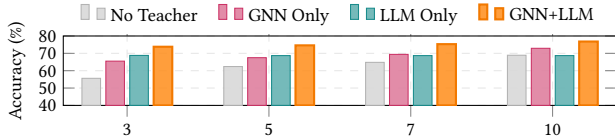

\begin{figure}[t]
\footnotesize
\centering
\begin{tikzpicture}
\begin{axis}[
    ybar,
    width=1.05\linewidth,
    height=2.5cm,
    bar width=6pt,
    ylabel={\# LLM Tokens},
    symbolic x coords={3,5,7,10},
    xtick=data,
    ymode=log,
    log basis y=10,
    ymin=50,
    ymax=20000,
    enlarge x limits=0.20,
    grid=major,
    grid style={gray!20,dashed},
    legend style={
        at={(0.5,1.02)},
        anchor=south,
        legend columns=4,
        draw=none,
        font=\footnotesize,
        /tikz/every even column/.append style={column sep=0.4cm}
    },
]

\addplot[
    fill=gray!40,
    draw=gray!70,
]
coordinates {
    (3,11701)
    (5,11701)
    (7,11701)
    (10,11701)
};
\addlegendentry{TAPE}

\addplot[
    fill=purple!50,
    draw=purple!80,
]
coordinates {
    (3,1440)
    (5,1440)
    (7,1440)
    (10,1440)
};
\addlegendentry{LLMGNN}

\addplot[
    fill=teal!60,
    draw=teal!90,
]
coordinates {
    (3,100)
    (5,100)
    (7,100)
    (10,100)
};
\addlegendentry{LLM4NG}

\addplot[
    fill=orange!85,
    draw=orange!95!black,
    line width=0.8pt
]
coordinates {
    (3,209)
    (5,191)
    (7,173)
    (10,173)
};
\addlegendentry{{\cote}}

\end{axis}
\end{tikzpicture}
\vspace{-5mm}
\caption{
Comparison of LLM token consumption under different few-shot settings (log scale).
}
\label{fig_3}
\vspace{-4mm}
\end{figure}
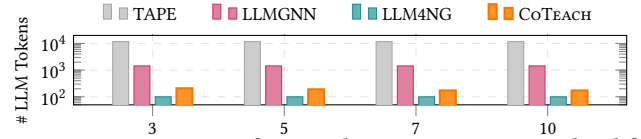

\vspace{1mm}
\noindent \textbf{EQ3: Cost-Effectiveness.} 
We further evaluate the cost-efficiency of {\cote} through LLM token consumption. 
As shown in Figure~\ref{fig_3}, {\cote} requires substantially fewer tokens than TAPE and LLMGNN, reducing token usage by up to 67.6× and 8.3×, respectively.
Although LLM4NG is more token-efficient, {\cote} achieves the best average rank, demonstrating a favorable cost-effectiveness trade-off.

\vspace{1mm}
\noindent \textbf{EQ4: Hyperparameter Sensitivity.} 
Lastly, we conduct sensitivity analyses on four key hyperparameters, including the loss weight ($\alpha$), GNN confidence threshold ($\gamma_g$), LLM confidence threshold ($\gamma_l$), and the ratio of selected seed nodes. 
The results show that {\cote} is generally robust to hyperparameter choices, with the best performance achieved under moderate confidence thresholds and sufficient seed node coverage. 
Detailed experimental results and analyses are provided at https://buly.kr/GP4yN34.
\section{Conclusions}~\label{s6}
In this paper, we propose {\cote}, a confidence-aware dual-teacher learning framework for few-shot text-attributed graph learning. 
Motivated by the observation that the reliability of LLM-derived supervision varies across nodes, {\cote} adaptively selects supervisory signals from GNN and LLM teachers based on prediction confidence. 
Experimental results on 4 real-world text-attributed graphs demonstrate that {\cote} achieves the best overall performance across most settings, validating the effectiveness of confidence-aware teacher selection. 
As future work, we plan to extend {\cote} beyond node-level tasks to edge-level and graph-level learning scenarios.


\begin{acks}
This work was supported by the National Research Foundation of Korea(NRF) grant funded by the Korea government(MSIT and MOE) (RS-2026-25494906 and RS-2025-2543583).

\end{acks}

\newpage

\section*{GenAI Usage Disclosure}~\label{s5}
In accordance with the ACM Authorship Policy, we disclose that generative AI tools were selectively employed during this research. Generative AI models (including ChatGPT-3.5 Turbo) were utilized as core components of the proposed LLM teacher in {\cote}. Beyond this intended methodological use, these models were not employed in any other stages of the research process, including data collection, preprocessing, experimental evaluation, or result analysis. GenAI-assisted writing support was strictly limited to editing and polishing author-written content, including minor grammatical corrections, phrasing refinements, and word-level autocorrections.

\bibliographystyle{ACM-Reference-Format}
\balance
\bibliography{sample-base}










\end{document}